\documentclass{article}

\usepackage{verbatim} 
\usepackage{amsmath,amssymb,amsfonts}
\usepackage{algorithmic}
\usepackage{algorithm}
\usepackage{multirow}
\usepackage{lipsum}
\usepackage{listings}
\usepackage{graphicx}
\usepackage{hyperref}
\usepackage{microtype}

\usepackage{subfigure}
\usepackage{booktabs} 
\usepackage{hyperref}
\usepackage{arxiv}

\usepackage[utf8]{inputenc} 
\usepackage[T1]{fontenc}    
\usepackage{url}            
\usepackage{booktabs}       
\usepackage{amsfonts}       
\usepackage{nicefrac}       
\usepackage{microtype}      
\usepackage{lipsum}

\title{ATL: Autonomous Knowledge Transfer from Many Streaming Processes\thanks{This paper has been accepted for publication in CIKM 2019. The source code is available in \url{https://bit.ly/2JS3F1u}.}}

\author{
	Mahardhika Pratama \\
	School of Computer Science and Engineering\\
	Nanyang Technological University, Singapore \\
	\texttt{mpratama@ntu.edu.sg} \\
	\And
	Marcus de Carvalho \\
	School of Computer Science and Engineering\\
	Nanyang Technological University, Singapore \\
	\texttt{marcus.decarvalho@ntu.edu.sg} \\
	\And
	Renchunzi Xie \\
	School of Computer Science and Engineering\\
	Nanyang Technological University, Singapore \\
	\texttt{enchunzi.xie@ntu.edu.sg} \\
	\And
	Edwin Lughofer\\
	Johannes Kepler University\\
	Linz, Austria \\
	\texttt{edwin.lughofer@jku.at} \\
	\And
	Jie Lu\\
	University of Technology Sydney\\
	Sydney,NSW,Australia \\
	\texttt{jie.lu@uts.edu.sg} \\
}

\begin{document}
	\maketitle
	
	\begin{abstract}
		Transferring knowledge across many streaming processes remains an uncharted territory in the existing literature and features unique characteristics: no labelled instance of the target domain, covariate shift of source and target domain, different period of drifts in the source and target domains. Autonomous transfer learning (ATL) is proposed in this paper as a flexible deep learning approach for the online unsupervised transfer learning problem across many streaming processes. ATL offers an online domain adaptation strategy via the generative and discriminative phases coupled with the KL divergence based optimization strategy to produce a domain invariant network while putting forward an elastic network structure. It automatically evolves its network structure from scratch with/without the presence of ground truth to overcome independent concept drifts in the source and target domain. Rigorous numerical evaluation has been conducted along with comparison against recently published works. ATL demonstrates improved performance while showing significantly faster training speed than its counterparts. 
	\end{abstract}


	\section{Introduction}
	The transfer learning problem across several streaming processes refers to prominent real-world problem where one is confronted by several streaming data processes following different data distributions. In realm of condition monitoring problems, several machines are used to manufacture different parts of aircraft engine and operate without interruption thus generating streaming environments. Input features are produced by performing a particular feature extraction method from the same signal references: force, vibration or acoustic emission. A predictive model can be built upon a single machine where true class label is usually fed back by operator from visual inspection of surface roughness. Nevertheless, building a machine-specific predictive model is cumbersome and costly because labelling process carries the cost of complete shutdown of overall machining process. This practical case leads to three challenges: 1) \textbf{scarcity of labelled samples}; 2) \textbf{covariate shift}; 3) \textbf{asynchronous drift}. The first problem is obvious because of the requirement of visual inspection of tool condition in the labelling process. The second problem occurs because data samples are drawn from the same feature space but different distributions. The last problem happens because different processes suffer from independent drifts due to aging of particular components, etc.   
	
	This problem goes beyond the major focus of existing literature which deals with a single data stream process. This bottleneck often forces one to treat multistream problem as single data stream problem with non-stationary property \cite{MSC}. This approach leads to the situation of unnecessary drift triggering due to the covariate shift property across different domains and lacks of physical interpretation because concept drift does not necessarily explain changing phenomena in a specific process such as aging components, external disturbances, or different operating conditions. Moreover, this technique leaves expensive labelling process unanswered and causes late detection of drift. Note that the multistream problem differs from conventional domain adaptation and transfer learning because the source and target processes continuously run in parallel and suffer from their own concept changes.  
	
	This problem, defined as multistream mining problem, has started to attract research attention. These approaches share commonalities where they apply the bias correction mechanism using the estimation of probability density ratio applied to weight the target domain such that the gap between two domains is minimized. Multistream classification (MSC) \cite{MSC} utilizes the kernel mean matching (KMM) method to perform estimation of probability density ratio. The MSC makes use of the ensemble paradigm where a new classifier is added if drift is detected in either domain. This approach incurs considerable structural complexity while the KMM method is re-executed if a drift is detected and imposes prohibitive computational burden - cubic time complexity \cite{KMM}. FUSION is proposed to remedy the drawback of MSC \cite{FUSION}. Unlike MSC, FUSION has inherent capability to address asynchronous drift directly measured from the density ratio itself. Unlike MSC, FUSION implements Kullback Leibler Importance Estimation Procedure (KLIEP) \cite{KLIEP} to perform the density ratio estimation and in turn to resolve the issue of sample bias. The KLIEP is, however, computationally expensive because its optimization procedure imposes quadratic computational complexity. MSCRDR utilizes the relative density ratio minimizing the Pearson's divergence \cite{PEARSON}. This approach defines the relative density ratio as the proportion of source density and weighted target density where the weight of target density is tuned using the expectation minimization principle. The three approaches are all built upon the non deep learning approach which might not scale well to handle high input dimension. 
	
	Autonomous transfer learning (ATL) is proposed as a deep learning solution of the transfer learning problem across several streaming processes. ATL possesses a flexible network structure where its network structure is self-evolving from data streams both in the supervised and unsupervised learning mode. This mechanism enables for effective handling of concept drifts both in the source and target domains. ATL is capable of coping with the multistream problem where source and target domain suffers from the sample bias problem with the absence of ground truth in the target domain. 
	
	Unlike conventional multistream analytic, ATL applies shared representation between source and target domain to address \textbf{the sample bias problem} between source and target domains. ATL works under the absence of any labelled samples in the target domain and the domain adaptation process is formulated as the joint optimization problem of three objectives: minimization of predictive error in the source domain, reconstruction error in the target domain, the probabilistic distance between source and target domains. 
	The self-evolving mechanism of network width is developed to address \textbf{the asynchronous drift problem}. That is, ATL is capable of adding and pruning its hidden units \textbf{from scratch} in both discriminative (drift in the source domain) and generative (drift in the target domain) phases. The self-organizing network structure is carried out by extending the network significance method \cite{ADL} suffering from unrealistic assumption of normal distribution and lack of adaptation to concept drifts. The NS method is remedied using the idea of autonomous Gaussian mixture model (AGMM) estimating the complex probability density function in both source and target domain. AGMM characterizes an open structure with growing, pruning and tuning mechanism thus following any variation of source and target data streams. This trait leads to accurate approximation of probability density function which navigates the NS method under the presence of the concept drift identification. Moreover, direct addition of $M$ hidden units instead of one by one addition is devised to expedite model's convergence where $M$ denotes the number of Gaussian components. This strategy is made possible due to the self-evolving nature of AGMM which explores the true complexity of data distributions. 
	
	The underlying contribution of this paper can be seen in three aspects: 1) this paper proposes novel multistream classification approach termed ATL using a single network structure in lieu of ensemble method as realized in the conventional multistream algorithms; 2) ATL characterizes an elastic network structure which can be automatically generated and pruned based on predictive error and reconstruction error. This trait addresses the asynchronous drift issue in the source or target domain while enhances the network significance method \cite{ADL} via solution to unrealistic assumption of normal distribution. It paves a way for direct addition of multiple hidden units thus speeding up model updates; 3) This paper also proposes the autonomous Gaussian mixture model (AGMM) performing the flexible estimation of probability density function. This approach addresses the unrealistic assumption of normal distribution and adapts to the concept drift causing the previous density estimation to be obsolete. Our approach also go one step ahead with direct addition of $M$ nodes at once which expedites to arrive at desirable level of network capacity. Our numerical evaluation and comparative study have confirmed the advantage of ATL where it outperforms prominent multistream classification or transfer learning algorithms in the aspect of predictive accuracy while being significantly faster than recently published multistream learners.
	\begin{figure}[t!]
		\begin{centering}
			\includegraphics[scale=0.4]{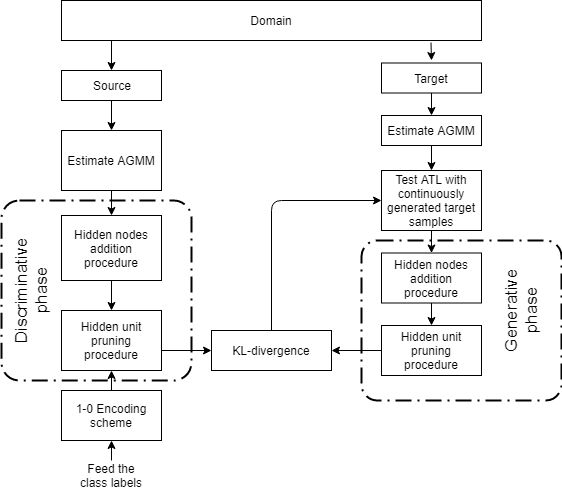}
			\par\end{centering}
		\caption{Learning Policy of ATL}
		\label{fig:ATL_policy}
	\end{figure}
	\section{Problem Formulation}
	The transfer learning problem in several streaming processes portrays two independent data generation processes $S,T$ running simultaneously and generating two data streams, namely source stream $B_1,B_2,...,B_{K_s}$ and target stream $B_1,B_2,...,B_{K_t}$ where $K_s,K_t$ are respectively the number of source and target streams often unknown in practise. We limit our discussion here for single source and target streams but the result can be continued for more complex case of multiple source and target domains. The source and target domains $S,T$ are drawn from the same feature and target spaces $B_{k_s}\in \Re, B_{k_t}\in \Re$ but are controlled by two different distributions $P_S(x)\neq P_T(x)$. In other words, \textbf{the covariate shift issue} occurs in both domain. The access of ground truth is granted only in the source domain $B_{k_s}=(X_S,Y_S)\in Re^{N_s\times(u+m)}$ where $u,m$ respectively stand for the number of input and output dimensions while $N_s$ denotes the size of source streams $N_s\geq1$. Note that the one-hot encoding scheme is applied to the target class label of source data samples. The target domain suffers from the absence of true class label $B_{k_t}=(X_T)\in\Re^{N_T\times u}$ where $N_T$ is the size of target stream. As the nature of data streams, both source and target streams are subject to their own concept drift $P_S(Y|X)_t\neq P_S(Y|X)_{t-1}$ and $P_T(Y|X)_t\neq P_T(Y|X)_{t-1}$ which may ensue in different time instants. In a nutshell, the underlying goal of multistream learner is to predict target data streams with the absence of ground truth but is supported by the labelled source data streams while adapting to any concept changes in both domains. In practise, a model is forced to perform the prediction first before performing model update due to possible delay in labelling data streams. \textbf{The prequential test-then-train} protocol is simulated in our numerical study. 
	
	\section{ATL: Autonomous Transfer Learning}
	An overview of ATL learning policy is visualized in Fig. \ref{fig:ATL_policy}. ATL learning algorithm consists of two learning stages: \textit{domain adaptation mechanism} and \textit{network width adjustment mechanism}. The learning process starts from the AGMM training phase where two AGMMs are created for both source and target domains in unsupervised manner. The discriminative phase takes place using labelled source domain samples and is equipped with the adjustment of network width facilitating the node growing and pruning process based on the classification error with the help of source AGMM. The generative phase in the target domain takes place afterward where the target AGMM is incorporated in the adjustment of network width via the bias (growing) and variance (pruning) estimation with respect to reconstruction error navigated with the target AGMM. Note that the adjustment of network width in both generative and discriminative phase using target and source AGMMs addresses the asynchronous drift problem. Furthermore, AGMM also has inherent drift handling capability where it is capable of automatically growing and pruning Gaussian's components, thereby leading to flexible estimation of probability density function $p(x)$. The last phase is the parameter learning phase of KL divergence. %
	
	ATL is structured under a single layer network with a single softmax layer connected for the discriminative phase while it is framed under a single layer denoising autoencoder (DAE) in the generative phase. Note that the transition of generative and discriminative phases occur in the closed loop fashion regardless of the domain of interest. That is, the two phases exploit shared parameters and are meant to produce a domain invariant network. Suppose that $R$ stands for the number of hidden nodes, the output of ATL is expressed as follows:
	\begin{equation}
	\hat{y}=softmax(s(X_n W_{in}+b)W_{out}+c)
	\end{equation}
	where $X_{n}\in\Re^{u},W_{in}\in\Re^{u\times R},b\in\Re^{R}$ respectively stand for the input vector of interest at the $n-th$ time instant, the input weight and bias while $W_{out}\in\Re^{R\times m},c\in\Re^{m}$ denote the output weight and bias of softmax layer. The sigmoid function is selected as the activation function. In the generative phase, ATL is formed under a single hidden layer denoising autoencoder (DAE) as follows:
	\begin{equation}
	f_{enc}=s(\tilde{X}_n W_{enc}+b);\hat{X}_n=f_{dec}=s(f_{dec} W_{dec}+d)
	\end{equation}
	where $\hat{X}_n$ is the reconstructed input features while $W_{dec}\in\Re^{R\times u},d\in\Re^{u}$ are the connective weight and bias of decoder. $W_{enc}\in\Re^{u\times R},b\in\Re^{R}$ are the input weight and bias of encoder and akin to the input weight and bias of discriminative network $W_{enc}=W_{in},b=b$. The tied-weight property of autoencoder is excluded here and the greedy layer-wise approach is applied to adjust $W_{enc},W_{dec}b,d$ respectively. The greedy layer-wise approach leads to better reconstruction power than that of the tied weight because it works without any constraint. Furthermore, $\tilde{X}_n$ stands for the corrupted input feature applying the masking noise where $n'$ of input attributes are set to zero. The DAE reconstructs its original clean feature space through the latent space producing representation of target domains and underpinning the knowledge transfer mechanism. Moreover, $W_{in},b$ are shared in both generative and discriminative phases in handling both source and target domain. The noise injecting mechanism is meant to extract robust features being stable against external perturbation and applied in the target generative phases. The main goal of generative phase is to recover original clean input representation from corrupted input space. Learning policy of ATL is pictorially exhibited in Fig. \ref{fig:ATL_policy} while the pseudocodes are provided in the supplemental document. 
	
	\subsection{Solution of Covariate Shift}
	The covariate shift problem is resolved in the parameter learning phase of ATL formed under a combined optimization problem:
	\begin{equation}\label{costfunction}
	L_{overall}=L(\hat{y}_{s},y_{s})+L(\hat{x}_{t},x_{t})+\Xi(h_{s},h_{t})
	\end{equation}
	where $L(\hat{y}_s,y_s)$ is the discriminative loss function in the source domain and $L(\hat{x}_t,x_t)$ is the generative loss function in the target domain. $\Xi(h_{d}^{s},h_{d}^{t})$ is the KL divergence loss function which minimizes the probabilistic distance of the source and target domains where $h_{s},h_{t}$ stand for the hidden mapping of ATL's network given the source and target data samples $x_s,x_t$. The intuition behind (\ref{costfunction}) can be understood in three facets: 1) ATL should classify labelled source data streams well because the access of ground truth is only available from the source domain; 2) the representation of source domain produced by the discriminative phase is further adapted to the target domain with the absence of ground truth under the generative phase. This step is an attempt to construct an approximation of ideal discriminative representation \cite{DRCN}; 3) The KL divergence loss aims to minimize the probabilistic distance between source and target domains. 
	
	This approach is inspired by \cite{TLDA}, \cite{DRCN} where the domain adaptation is set as the generative and discriminative learning problems. Our approach distinguishes itself in its application into the online context as well as the aspect of evolving data streams in which the source and target domains characterize streaming data problems affected by the concept drifts in both domains in different activation regions and time intervals. This task is handled using a self-evolving network.  Furthermore, the concept of DAE is applied here in lieu of AE strategy which implements the noise injecting mechanism as the regularization strategy to address the overfitting issue in the generative phase. In other words, no explicit regularization term is applied in (\ref{costfunction}) and the regularization task is taken over by partially destroying input features. As explained in \cite{learning_dynamic_AE}, the noise injecting mechanism allows a model to converge faster than using the regularization term provided the selection of correct noise level. Because the network parameters are shared across the three optimization stages and the learning process is played in the closed loop fashion, no additional regularization term is integrated in the discriminative phase as well as the KL divergence phase. Furthermore, the use of tradeoff parameters usually applied as the weighting factor of each cost function \cite{TLDA,DRCN} is removed here because these parameters are usually problem-dependent and hard to select without rigorous pre-training phase. 
	
	(\ref{costfunction}) is solved via alternate optimization approach. The optimization problem makes use of the stochastic gradient descent approach with \textbf{no epoch} or \textbf{epoch per mini batch}. That is, a sample or mini batch is not revisited again once learned. This step aims to demonstrate the feasibility of ATL for deployment in the most challenging situation of data stream learning. The minibatch here is interpreted where the iteration over a number of epoch is carried out in a minibatch without revisiting it again in the future thus still retaining bounded computational and memory demand. Furthermore, the structural learning phase takes place in both generative and discriminative phases in a closed loop fashion and in both domains. The network structure of the generative phase is fed as an initial point of the discriminative phase.     
	\textbf{Discriminative Learning Phase:} the labelled streaming data $B_{k_{s}}=(X_S,Y_S)$ of the source domain is learned in the supervised fashion using the stochastic gradient descent method. The discriminative phase is carried out by minimizing the negative entropy function:
	\begin{equation}\label{discriminative}
	\min_{W_{in},W_{out},b,c}-\frac{\sum_{n=1}^{N_s}\sum_{o=1}^{m}1(o=y_{n})log(\hat{y}_{n})}{N_s}
	\end{equation}
	where $\hat{y}_n$ is the output of softmax layer while $y_n$ is the target vector which implements the one-hot encoding scheme. The term $1(.)$ is only true in respect to the $1$ element of target vector as a result of the one-hot encoding scheme. 
	Generative Learning Phase: the generative training process performs the representation learning mechanism via the encoding and decoding mechanism of denoising autoencoder (DAE) \cite{VincentDAE}. That is, the underlying objective is to extract useful representation of both domains by minimizing their squared reconstruction error $(\hat{x}-x)^{2}$. The cost function is formulated as follows:
	\begin{equation}
	L(\hat{x},x)=\sum_{n=1}^{N_{t}}\frac{1}{N_{t}}(x_{n}^{t}-\hat{x}_{n}^{t})^{2}
	\end{equation}
	where the generative training process is undertaken in the target domain to generate an ideal discriminative representation of the target domain. That is, the input space is partially destroyed by noise where $u'$ elements of input variables $\tilde{x}$ are set to zero in random fashion while the target variable is the clean input attributes themselves $x$. Note that the noise injecting mechanism here functions similarly as the regularization and in fact converges faster than regularization provided that noise is provided at a right level \cite{learning_dynamic_AE}. The structural learning phase occurs in both generative and discriminative phase meaning that the generative phase is capable of addressing the covariate drift in the target domain while the discriminative phase adapts to the real drift of the target domain by having the access to true class label. Since the encoding parameters $W_{in},b$ are shared in both generative and discriminative phases, the generative phase of the target domain generates the target domain representation guiding the softmax layer of source domain to predict target domain samples equally well as those of source domain samples. 
	KL Divergence: The optimization of KL divergence cost function targets minimization of two probability distributions between the source and target domains. The KL divergence, known as the relative entropy, quantifies the divergence of two distributions and is applied independently in the hidden layer of ATL, $h^{S},h^{T}$ aligning the activation degrees of ATL under the source and target domains.  $h^{s},h^{t}\in\Re^{R}$ are respectively the activation degrees of the hidden layer given the source domain samples $x_s$ and the target domain samples $x_t$. The cost function, $\Xi(h^{s},h^{t})$, is written as follows:
	\begin{equation}\label{KL}
	\Xi(h^{s},h^{t})=KL(h^{s}||h^{t})+KL(h^{t}||h^{s})
	\end{equation}
	\begin{equation}
	KL(h^{s}||h^{t})=\sum_{i}^{R}\Pi_{i}^{s}\ln(\frac{\Pi_{i}^{s}}{\Pi_{i}^{t}})
	\end{equation}
	\begin{equation}
	\pi_{i}^{s,t}=\frac{1}{n_{s,t}}\sum_{n=1}^{n_{s,t}}h_{i,n}^{s,t};
	\Pi_{i}^{s,t}=\frac{\pi_{i}^{s,t}}{\sum_{i=1}^{R}{\pi_{i}^{s,t}}} 
	\end{equation}
	This formula estimates the information lost when describing $h^{s}$ through $h^{t}$ and the dissimilarity is inversely proportional to the KL index $KL(h^{s}||h^{t})$. The underlying objective of the KL-based optimization problem is to suppress the difference between the two probability distributions and to align both domains such that prediction of target domain samples is possible without seeing their true class labels during the traning phase. This domain adaptation strategy aims to produce a domain invariant network which describes both source and target domains equally well.
	\begin{figure*}[t!]
		\begin{centering}
			\includegraphics[scale=0.4]{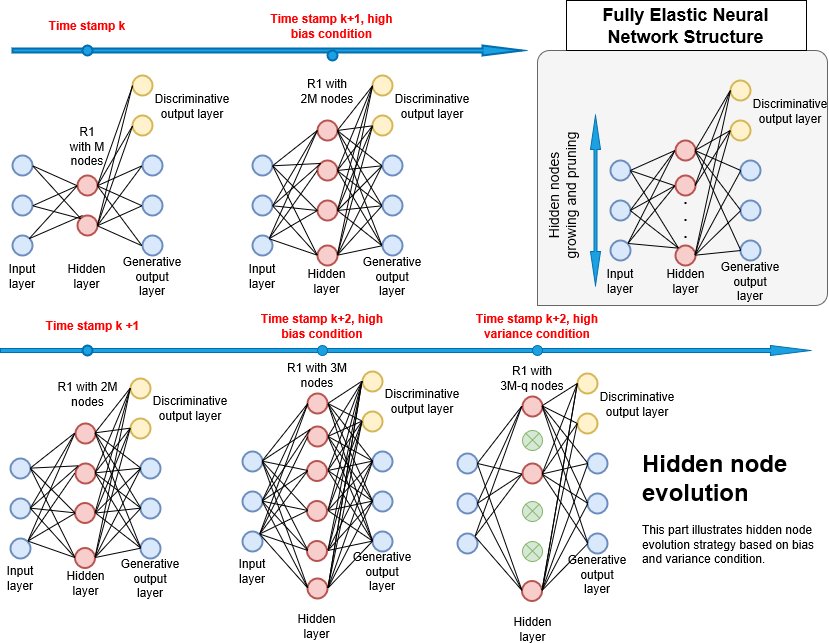}
			\par\end{centering}
		\caption{ATL's structural evolution.}
		\label{fig:Network Evolution}
	\end{figure*}
	\subsection{Evolution of Network Width}
	The elastic network width of ATL is steered by the idea of bias and variance decomposition advancing the network significance (NS) method in \cite{ADL}. It uses the autonomous gaussian mixture model (AGMM) to perform estimation of complex probability density function $p(x)$. The use of AGMM resolves strict assumption to normal distribution as occurred in the NS method while characterizes an open structure via the growing and pruning mechanism of Gaussian's components instead of conventional GMM. The flexible nature of AGMM improves the adaptive mechanism of ATL against the concept drift because distributional change of data streams leads current estimation of probability density function to be obsolete $p(x)_{t}\neq p(x)_{t+1}$  and is handled by introducing new Gaussian components on demand. The use of AGMM makes possible to insert a dynamic number of hidden units $M$ during the node growing phase where $M$ is the number of Gaussian components.  
	
	\textbf{Hidden Unit Growing Strategy}: The hidden unit growing strategy of ATL follows the fact that the network capacity should increase in the under-fitting situation, \textbf{high bias condition} which mirrors the model's generalization power. That is, the generalization error is derived from the expectation of squared error given a particular probability density function $p(x)$ or also defined as the network significance measure in \cite{ADL} $NS=\int_{-\infty}^{\infty}(y-\hat{y})p(x)dx$. This expression leads to the popular bias-variance decomposition $NS=(E[\hat{y}]-y)^{2}+(E[\hat{y}^{2}]-E[\hat{y}]^{2})=Bias^{2}+Var$. The estimation of network bias depends on the expectation of network output $E[\hat{y}]=\int_{-\infty}^{\infty}s(X_{n}W_{in}+b)W_{out}p(x)dx+c$. The underlying bottleneck of original NS method lies in the strong assumption of normal distribution $p(x)=N(x;\mu,\sigma)$. In addition, the use of a single Gaussian component does not fit the nature of concept drifts undermining its relevance to capture new data patterns $p(x)_{t}\neq p(x)_{t+1}$. 
	
	The aforementioned problem is addressed here using the concept of AGMM where $M$ Gaussian components are self-evolved to inform current data trend $p(x)=\sum_{m=1}^{M}N(x;\omega_m,\mu_m,\sigma_m)$ where $\mu_m,\sigma_m,\omega_m$ are respectively the center of $m-th$ Gaussian component, the width of $m-th$ Gaussian component and the mixing coefficient of $m-th$ Gaussian component. By following the result in \cite{Murphy_Machine_Learning} where the sigmoid function can be approximated with the probit function and the integral of probit function is in fact another probit function, the expectation of network output is expressed:
	\begin{equation}\label{expectation}
	E[\hat{y}]=\sum_{m=1}^{M}\omega_m s(\frac{\mu_m}{\sqrt{1+\pi\frac{\sigma_m^{2}}{8}}}W_{in}+b)W_{out}+c
	\end{equation}
	The network bias expression can be established with ease based on (\ref{expectation}). (\ref{expectation}) can be generalized for the $D$ network depth by simply applying the forward propagation approach. Note that the number of Gaussian components $M$ is fully data driven rather than fixed. A high bias condition triggering the hidden node growing process is determined using an adaptive version of the $\chi$ sigma rule \cite{GamaDataStream}: 
	\begin{equation}\label{Grow}
	\mu_{Bias}^{t}+\sigma_{Bias}^{t}\geq\mu_{Bias}^{min}+\chi\sigma_{Bias}^{min}
	\end{equation}
	where $\chi=1.25\exp{-Bias^{2}}+0.75\in[0.75,2]$ leading to a dynamic confidence level of the $\chi$ sigma rule in the range of $60.8\%$ and $95.5\%$. This strategy overcomes the introduction of user-defined parameter $\chi$ well-known to be problem-specific. That is, introduction of new hidden units is forced to occur in the high bias case whereas it is avoided in the low-bias condition to prevent the problem of over-fitting. Furthermore, $\mu_{Bias}^{min},\sigma_{Bias}^{min}$ are reset whenever (\ref{Grow}) is observed whereas no reset mechanism is applied to $\mu_{Bias}^{t},\sigma_{Bias}^{t}$ because of the nature of network bias which should be examined across all patterns. Our empirical study also found that the hidden units grows uncontrollably when resetting $\mu_{Bias}^{t},\sigma_{Bias}^{t}$. The node growing mechanism is carried out in both generative and discriminative fashions where the underlying difference lies only in the use of squared reconstruction error for the generative step while the discriminative step exploits the squared predictive error. If (\ref{Grow}) is come across, $M$ number of hidden units are inserted at once $R=R+M$ and initialized by following Xavier initialization. This strategy aims to expedite the growth of network structure to attain a desirable level of network capacity. Note that this strategy goes one step ahead of incremental addition of hidden units \cite{ADL} often ending up with a too few number of hidden units.   
	
	\textbf{Hidden Unit Pruning Strategy}: The hidden unit pruning mechanism follows the same principle of the hidden unit growing strategy. It, however, distinguishes itself from the hidden unit growing strategy in the application of network variance rather than the network bias which accurately signifies the overfitting situation. The network variance $Var=E[\hat{y}^{2}]-E[\hat{y}]^{2}$ can be established by exploiting (\ref{expectation}). Note that the expectation $E[\hat{y}^{2}]$ is the I.I.D variable leading to $E[\hat{y}^{2}]=E[s(X_{n} W_{in}+b)]E[s(X_{n} W_{in}+b)]W_{out}+c$. 
	
	Hidden units are discarded in the high variance situation formulated using the $\gamma$ sigma rule. The $\gamma$ sigma rule aims to capture abnormal situations presenting distinct case of regular observation:
	\begin{equation}\label{pruning}
	\mu_{Var}^{t}+\sigma_{Var}^{t}\geq\mu_{Var}^{min}+2*\gamma\sigma_{Var}^{min}
	\end{equation}
	where $\gamma=1.25\exp{(-Var^{2})}+0.75\in[0.75,2]$ leading to a flexible $\gamma$ sigma rule which adapts to the network variance hovering around $\gamma=[0.75,2]$ and supports for flexible confident levels in the range of $[60.8\%,99.9\%]$. That is, hidden unit pruning strategy occurs frequently in the high variance case whereas it is precluded in the low variance situation. As with the hidden unit growing mechanism, $\mu_{Var}^{min},\sigma_{Var}^{min}$ are reset if (\ref{pruning}) is satisfied and function in both generative and discriminative fashion. Furthermore, the term $2$ is included in (\ref{pruning}) to avoid the direct-pruning-after-adding case because initially introduction of new nodes causes temporary increases of network variance but it slowly decreases as incoming observations are encountered. 
	
	The hidden unit pruning mechanism is activated once observing (\ref{pruning}). This mechanism examines the statistical contribution of hidden units and inconsequential nodes are defined as those playing little given the AGMM-based probability density function $p(x)$. The statistical contribution of a hidden node is derived as the expectation of the activation function as follows: 
	\begin{equation}
	HC_i=\int_{-\infty}^{\infty}s(X_n W_{in}+b)p(x)dx=\sum_{m=1}^{M} s(\frac{\mu_m}{\sqrt{1+\pi\frac{\sigma_m^{2}}{8}}}W_{in}+b)  
	\end{equation}
	A hidden unit is subject to the pruning process if its statistical contribution falls below the average statistical contribution of hidden units as follows:
	\begin{equation}\label{pruning}
	HC_i<\mu_{HC}-\sigma_{HC}
	\end{equation}
	where (\ref{pruning}) opens opportunity for several hidden nodes to be gotten rid of at once. This strategy aims to induce rapid complexity reduction mechanism which counterbalances the direct addition of $M$ hidden units in the hidden unit growing mechanism. 
	
	\textbf{Autonomous Gaussian Mixture Models}: The underlying innovation of ATL's hidden unit growing and pruning mechanisms lies in the deployment of autonomous gaussian mixture model (AGMM) generating the complex probability density function $p(x)$. AGMM characterizes self-evolving nature where it is equipped by the growing and pruning mechanism tracking any variation of data streams. The growing process of AGMM relies on the autonomous clustering approach which exploits spatial proximity of incoming sample to existing Gaussian components. A new Gaussian component is added if a data sample is sufficiently distinct and beyond the zone of influence of existing Gaussian components. This measure is realized with a threshold-free compatibility test as follows:
	\begin{equation}\label{threshold}
	\max_{m=1,..,M}\min_{j=1,..,u} exp(-\frac{(x-\mu_{j,m})}{2\sigma_{j,m}^{2}}^{2})<exp(-\frac{u*\chi}{4-2exp(-u/2)})
	\end{equation}
	The compatibility test examines the matching degree of incoming sample and is statistically sound because the right side of (\ref{threshold}) is derived from the sigma rule. That is, a data sample is considered as a novel sample if it occupies beyond $\mu_{j,m}+\chi\sigma_{j,m}$. As with (\ref{Grow}), $\chi$ is enumerated from the network bias to suit to the level of network bias while $u$ is included to incorporate the effect of input dimension. In addition, the $\min$ operator is used in lieu of dot product operator because it is robust against the curse of dimensionality where the matching degree rapidly decreases as the input dimension. Nonetheless, we found that the use of only (\ref{threshold}) in generating Gaussian components causes excessive number of Gaussian components mainly due to false alarms primarily associated with the low variance direction of data distributions. A vigilance test based on the concept of available space is proposed as another growing criterion and decides to what extent existing Gaussian components can be relied upon to capture variations of data streams. 
	
	A vigilance test is incorporated as another growing method of AGMM controlling the size of Gaussian components. A new Gaussian component is added only if the winning component having the closest relationship to an incoming sample does not have any available space to expand its size. That is, the tuning process of winning Gaussian component incurs significant overlapping to other Gaussian components. This strategy is also meant to address the cluster delamination problem where one Gaussian component covers several distinct concepts due to its over-sized coverage. This issue leads to too general Gaussian component which loses its relevance to portray particular concepts. The vigilance test is formulated:
	\begin{equation}\label{vigilance}
	\hat V_{win} \geq \rho\sum_{m=1}^{M}\hat V_m
	\end{equation}
	where $\rho$ stands for the control parameter. That is, it is determined from the current location of winning component whether it is overlapping or partially overlapping. Note that no overlapping case is irrelevant in this case due to the infinite support property of Gaussian function where the degree of membership is never exactly zero. An overlapping score $overlap_m$ is assigned and specifies a relationship of the Gaussian component to $m-th$ Gaussian component:
	\begin{equation}
	overlap_m=
	\begin{cases}
	1/(M-1)\longrightarrow full\\
	1/(M-1)*reward\longrightarrow partial\\
	\end{cases}
	\end{equation}
	where the full overlap case presents a situation where a Gaussian component $N(\mu_m,\sigma_m)$ is fully contained by the winning component $N(\mu_{win},\sigma_{win})$ while the partial overlap indicates one-sided overlapping case - one Gaussian is not contained by another. $reward$ is obtained from the normalized distance between the two components $\frac{||\mu_{m}-\mu_{win}||}{||\mu_{m}+\mu_{win}||}+\frac{||\sigma_{m}-\sigma_{win}||}{||\sigma_{m}+\sigma_{win}||}$. The score of non-overlapping case is simply arranged as zero. The degree of overlapping in the three cases are elicited from comparison between $\mu_{m}\pm \sigma_m$ and $\mu_{win}\pm \sigma_{win}$. This strategy assigns an overlapping score to other Gaussian components based on its relationship to the winning components. The more adjacent component the lower the score is allocated, thus pushing to the decision of addition of a new component. On the other side, the more distant component the higher score is received meaning that it permits the winning component to update its location and to enlarge its size. This strategy is in line with the concept of available space where a new component is added only if the update of current cluster compromises the generalization power because it no longer represents a particular concept - too general. The control parameter is finally calculated as $\rho=\sum_{m=1}^{M-1}overlap_m$ where its value does not go beyond the range of $[0.1,1]$ to guarantee reasonable size of the winning component. That is, its size cannot be smaller than 10\% the total volume of all components while it cannot be bigger than the total volume of all components.     
	
	A new component is incorporated $M=M+1$ if the two criteria in (\ref{threshold}) and (\ref{vigilance}) are satisfied. That is, a new component is crafted using the sample of interest as the center of the new component $\mu_{M+1}=X_n$ while the width of a new Gaussian component is set as its one-dimensional distance to the winning Gaussian component $\sigma_{M+1}=|X_n-C_{win}|$. In addition, $M$ is utilized as the hidden unit addition factor in which direct addition of $M$ nodes is implemented if (\ref{Grow}) is met. 
	
	The tuning case is applied to the winning component if either (\ref{threshold}) or (\ref{vigilance}) or both of them are violated. This condition observes the fact where the winning component is valid to embrace incoming sample or has likelihood to grow its zone of influence. The tuning strategy of Gaussian component is formalized \cite{BARTFIS}:
	\begin{equation}\label{GMMtuning}
	\mu_{j,win}=\mu_{j,win}+\frac{x-\mu_{j,win}}{Sup_{win}+1};
	\end{equation}
	\begin{equation}\label{GMMtuning1}
	\sigma_{j,win}^2=\sigma_{j,win}^2+\frac{(x-\mu_{j,win}^{2})-\sigma_{j,win}^{2}}{Sup_{win}+1};Sup_{win}=Sup_{win}+1
	\end{equation}
	where $Sup_{win}$ denotes the supports or population of the winning Gaussian component while it is incremented if the tuning condition is observed. The tuning strategy aims to move the center of Gaussian component close to the incoming sample while enlarging the influence zone of Gaussian component. This condition also guarantees the convergence property of (\ref{GMMtuning}) and (\ref{GMMtuning1}) as the increase of its population where a component having high population tends to be less responsive to accept new training stimuli than those of low populations. Forgetting mechanism can be implemented by reduction of component's supports. In addition, the winning Gaussian component is determined as that $N(\omega_{m},\mu_m,\sigma_m)$ having the closest distance to the sample of interest $X_n$.  
	
	Since AGMM governs the ATL network evolution, the compactness of AGMM plays key role to keep pace with high-pace data streams. The activity measure is utilized and delves the activity degree of Gaussian components over their lifespan. The degree of activity is defined as the accumulated matching degree of a Gaussian component since it was born up to the current time instant:
	\begin{equation}\label{AGMMprune}
	\phi_m=\frac{\sum_{i=1}^{Life_m}\min_{j=1,..,u}\exp{-1/2(x_j-\mu_{j,m})^{2}/\sigma_{j,m}^{2}}}{Life_m}  
	\end{equation}
	where $Life_m=N-born_m$ stands for the lifespan of $m-th$ Gaussian unit. $N$ denotes the number of data samples seen thus far while $Born_m$ labels a time index when $m-th$ Gaussian unit is generated. The pruning condition is arranged using the half-sigma rule $\phi_m\leq \mu_{\phi_{1:M}}-0.5*\sigma_{\phi_{1:M}}$. The pruning condition opens opportunity for more than one component to be removed at once $M=M-M_{red}$. $M_{red}$ denotes the number of Gaussian components captured by the half-sigma rule. Note that this procedure can be executed recursively without revisiting preceding samples. It is worth mentioning that for the implementation of (\ref{AGMMprune}) newly created components are exempted from the pruning process over an evaluation window. This strategy aims to give sufficient time period to observe the contribution of new components. The evaluation window is set as the size of data batch. 
	
	The mixing coefficient $\omega_m$ satisfies the partition of unity simply calculated using the posterior probability $P(N_m|X_n)$ expressing the probability of a sample to fall into the $m-th$ Gaussian unit:
	\begin{equation}\label{posterior}
	P(N_m|X_n)=\frac{P(X_n|N_m)P(N_m)}{\sum_{m=1}^{M}P(X_n|N_m)P(N_m)}
	\end{equation}
	\begin{equation}\label{likelihood}
	P(X_n|N_m)=\frac{1}{\sqrt{2\pi^{u}\hat{V}_m}}exp(-1/2\sum_{j=1}^{u}(x_n-\mu_{j,m})^{2}/\sigma_{j,m}^{2})
	\end{equation}
	\begin{equation}\label{prior}
	P(N_m)=\frac{Sup_m}{\sum_{m=1}^{M}Sup_m}
	\end{equation}
	where $P(X_n|N_m)$ stands for the likelihood function  while $P(N_m)$ denotes the prior probability. The mixing coefficient controls the influence of Gaussian components during the calculation of expected output (\ref{expectation}) and takes into account both the distance and population of Gaussian components. In addition, the term $\sqrt{2\pi\hat{V}_m}$ in (\ref{likelihood}) is often expensive to compute in the high dimensional problems and the volume of Gaussian component quickly approaches zero. It is approximated with $\sqrt{2\pi\min_{j=1,..,u}\sigma_{j,m}}$ in this case.  
	
	Two GMMs are created specifically for source and target domains in order for sample bias between the two domains to be carefully considered while its self-adaptive mechanism addresses independent drifts in both domains conveninetly. Note that this property also enhances the flexibility of node growing and pruning mechanisms (\ref{Grow}),(\ref{pruning}) hindered when applying a fixed number of Gaussian components. Because AGMM explores the orientation of data distributions, this trait makes possible to exploit their number of components as representation of problem complexity and in turn to answer a basic question of how many nodes to be added. Moreover, (\ref{Grow}), (\ref{pruning}) do have an aptitude to handle the concept drift because it adopts the idea of statistical process control (SPC) \cite{GamaDataStream}. It differs from \cite{GamaDataStream} in which it is not modelled as binomial distribution since it is derived from the network bias rather than accuracy matrix. Illustration of ATL's Network evolution is visualized in Fig. \ref{fig:Network Evolution} while the pseudo code is supplied in the supplemental document.   
	\begin{table}[t]
		\caption{Properties of the datasets.}
		\begin{centering}
			\label{properties} { 
				\begin{tabular}{lcccccc}
					\hline 
					\hline 
					& F & \#C & SS & TS & NC & Char \tabularnewline
					\hline 
					\hline 
					F. Covertype  & 54 & 7 & 291K & 291K & 582 & Stationary \tabularnewline
					HEPMASS 5\%   & 27 & 2 & 250K & 250K & 500 & Non-stationary     \tabularnewline
					Hyperplane    & 4  & 2 & 60K  & 60K & 120 & Non-Stationary \tabularnewline
					KDDCup        & 41 & 2 & 250K & 250K & 500 & Non-Stationary \tabularnewline
					SEA           & 3  & 2 & 50K  & 50K  & 100 & Non-Stationary \tabularnewline
					SUSY          & 18 & 2 & 2.5M & 2.5M & 5000 & Stationary \tabularnewline
					Weather       & 3  & 2 & 9K   & 9K & 18 & Non-Stationary     \tabularnewline
					\hline 
					\hline 
			\end{tabular}} 
			\par\end{centering}
		\centering{F: Features, C: Classes, SS: Source Samples, TS: Target Samples, Char: Characteristics}
	\end{table}
	\section{Numerical Evaluation}
	The efficacy of ATL is thoroughly examined under four evaluation procedures: 1) our numerical study starts with performance evaluation of ATL under the unsupervised transfer learning setting. The source and target samples are drawn from the same domains but suffers from sample bias problem akin to that of \cite{FUSION,MSC} - Table \ref{result};  2) because ATL must still retain reliability in the source domain, the performance of ATL for the source domains is evaluated - Table \ref{result}; 3) Ablation study is carried out to understand the advantage of each learning modules - Table \ref{ablation_result}; 4) the last section is to study the effect of the number of epochs. Note that the iterative learning over a number of epoch here is carried out per mini batch or data chunk where a data chunk is completely discarded once learned - Table \ref{result}. Visualization of ATL's learning performance is offered in \textbf{the supplemental document.} 
	\subsection{Benchmark Problem}
	\textbf{Dataset}: seven prominent data streams problems are used in our simulation: ForestCover \cite{forestcover}, SEA \cite{SEA}, SUSY \cite{Baldi2014SearchingFE}, KDDCup \cite{KDDCup}, hyperplane \cite{Bifet07learningfrom}, Hepmass(5\%) \cite{Baldi2014SearchingFE}, and weather \cite{DitzlerImbalanced} in which all of them except SUSY and Hepmass(5\%) characterize non-stationary characteristics but the two problems have a large number of instances which allows to simulate real data stream environments. Table \ref{properties} outlines dataset properties and they are detailed as follows: 
	
	\textit{Forest cover problem} presents a binary classification problem where the underlying goal is to identify the actual forest cover type, 30 by 30 cells, based on the US Forest Service (USFS) Region 2 Resource Information System (RIS) data. 
	
	\textit{KDDCup problem} is an intrusion detection problem detecting an attack of network connection, binary classification problem. The non-stationary characteristics comes from simulation of different types of network intrusion in a military network environments. \textit{SUSY problem} is a popular big data problem where it predicts a signal process leading to super-symmetric particles from the Monte-Carlo simulation - binary classification. 
	
	\textit{SEA problem} is a binary classification problem describing $f_1+f_2<\theta$ where the non-stationary property is caused by variation of the threshold $\theta=4\longrightarrow7\longrightarrow4\longrightarrow7$ inducing the abrupt and recurring drift. 
	
	\textit{Weather Problem} is a weather prediction problem where the goal is to perform one-day-ahead weather prediction problem whether rain occurs, the binary classification problem. This problem characterizes seasonal change of weather condition as well as the long term climate change because it covers over 50 years weather data. 
	
	\textit{Hyperplane Problem} is a synthetic data stream problem generated by the MOA software framework. This problem is a binary classification problem based on the $d$ dimensional random hyperplane $\sum_{j=1}^{d}w_jx_j>w_0$. This problem features the gradual drift problem where initially data samples are drawn from one distribution slowly replaced with another distribution. It has a transition period where data are drawn from mixture of the two distributions. 
	
	\textit{Hepmass(5\%) Problem} is a popular big data problem which presents a binary classification problem about separation of particle-producing collisions from a background. Original Hepmass problem consists of over 10 million records but only 5\% of which are used here. Note that this reduction does not diminish validity of our experiment due to the stationary trait of the Hepmass problem.
	
	\textbf{Simulation Protocol}: all problems are simulated under the prequential test-then-train procedure where all data samples are divided into equal-sized data chunks without changing the order. The first chunk is created in the warm-up period while the rest of data chunks create pseudo-streaming environments. The testing phase is carried out first for each mini-batch before training process and is performed across all points of data batch. Numerical evaluation is undertaken in the four angles as specified in Table \ref{result}. Accuracy is calculated independently per batch as the guideline in \cite{datastreamevaluation} and presented in Table \ref{result} as the average of numerical results across all batches. This approach avoids the learning performance during the stationary period to over-dominate the numerical results. On the other hand, the hidden node is measured as the final number of hidden nodes generated during the training process to allows fair comparison with benchmarked algorithms having static structure. The same case is implemented for the training time.  
	
	The covariate shift problem is induced by following the same procedure as \cite{FUSION,MSC} where the mean of each mini batch is computed as a reference point $\hat{x}$. The probability of a point to be grouped as source data samples follows a normal distribution $exp(-\frac{||x-\hat{x}||^{2}}{\sigma})$ where $\sigma$ here is simply the standard deviation of the data chunk. $N_S$ data samples are selected for every chunk in respect to the normal distribution such that both domains are biased. The rest of samples, $N_t$, are chosen for the target samples accordingly. In a nutshell, our numerical study follows the same protocol as applied in \cite{FUSION,MSC}. 
	
	\textbf{Baselines}: ATL is compared against two prominent multistream classification algorithms: FUSION \cite{FUSION}, MSC \cite{MSC}, Transfer Learning with Deep Autoencoder (TLDA) \cite{TLDA} and autonomous deep learning (ADL) \cite{ADL} chosen because they are either recently published multistream algorithms or popular transfer learning algorithms developed from the autoencoder (AE) concept. Their characteristics are discussed as follows:
	\textit{MSC} \cite{MSC} is an ensemble-based multistream classification algorithm applying the kernel mean matching (KMM) method as the domain adaptation algorithm and standalone concept drift detection method in the source and target domain. \textit{FUSION} \cite{FUSION} is an improvement of MSC algorithm using the KLIEP as the domain adaptation strategy and the asynchronous drift is detected by directly analyzing the density ratio.
	\textit{TLDA} \cite{TLDA} is an autoencoder-based transfer learning algorithm which does not yet have the self-organizing mechanism. Its domain adaptation strategy relies on the combination of generative training in the both source and target domains, discriminative training in the source domain, and KL divergence. Note that ATL domain adaptation strategy makes use of noise injecting mechanism rather than the explicit regularization term and the generative training phase is carried out only in the source domain. 
	\textit{ADL} \cite{ADL} is a self-organizing deep neural networks having the drift handling capability through dynamic network depth and width. ADL here treats the source and target domain as a single stream where the training process is performed in the source domain while using the target domain samples as the testing phase. 
	
	Numerical results are produced by running their published codes in the same computational environments and the hyper-parameters of the four algorithms are selected as per their original setting as reported in their original papers for the sake of fair comparison. if they perform surprisingly poor, their hyper-parameters are tuned and their best-performing results are reported here. Because TLDA possesses a static network structure, its network structure is assigned such that it carries the same level of network capacity as ATL. TLDA is executed with 10 epochs as per its original setting due to its batch characteristic. 
	
	On the other hand, ATL makes use two hyper-parameters, namely learning rate and momentum rate of SGD method, blindly selected at fixed values 0.01, 0.95 respectively in all parts of our simulations. This setting is to demonstrate that ATL is applicable to a wide range of cases without necessity of going through laborious pre-training phase. Numerical results are reported in Table \ref{result} where ATL's numerical results are obtained from the average of five independent runs. \textbf{To ensure reproducibility of our works, the source code of ATL is provided in \footnote{\url{https://bit.ly/2JS3F1u}}.}   
	
	\begin{table}[!t]
		\caption{Numerical results of consolidated algorithms.}
		\footnotesize
		\begin{centering}
			\setlength{\tabcolsep}{2pt} 
			\renewcommand{\arraystretch}{1.3} 
			\label{result} \scalebox{1.1}{ 
				\begin{tabular}{ccccccc|ccc}
					\toprule
					\toprule
					&       &       &       &       &       &       & ATL   & ATL   & ATL \\
					Dataset &  Config     & FUSION & MSC   & TLDA  & ADL   & \textbf{ATL} & \textbf{3 eps} & \textbf{5 eps} & \textbf{10 eps} \\
					\midrule
					\midrule
					& TCR(\%) & $87.72$ & $81.75$ & $63.84$ & $89.90$ & $\textbf{91.12}$ & $92.38$  & $92.50$  & $\textbf{92.98}$ \\
					& SCR(\%) & NA      & NA               & NA      & $91.58$ & $\textbf{92.50}$ & $93.27$  & $93.53$  & $93.68$  \\
					SEA & TrT(s)  & $28777$  & $3848$               & $6309$   & $10$  & $262$    & $514$    & $824$    & $1570$    \\
					& HN      & NA      & NA               & $100$   & $36$     & $106$   & $110$    & $149$    & $157$    \\
					
					\midrule
					& TCR(\%) & $75.83$ & $\textbf{82.22}$ & $66.58$ & $68.69$ & $71.53$ & $71.97$  & $72.82$  & $72.09$  \tabularnewline
					& SCR(\%) & NA      & NA      & NA      & $70.10$ & $70.21$ & $69.45$  & $69.24$  & $69.58$ \tabularnewline
					Weather	& TrT(s)  & $6234$ & $148$  & $226$  & $1.81$    & $66$            & $169$    & $291$    & $618$           \tabularnewline
					& HN      & NA      & NA      & $100$   & $7$    & $112$            & $264$    & $365$    & $408$            \tabularnewline
					\hline 
					
					& TCR(\%) & $99.95$  & $99.91$ & $99.01*$      & $\textbf{99.99}$ & $99.52$ & $99.53$ & $99.53$ & $99.95$ \tabularnewline
					
					& SCR(\%) & NA       & NA      & NA      & $\textbf{99.68}$ & $92.61$ & $96.72$ & $98.34$ & $99.01$ \tabularnewline
					
					KDDCup	& TrT(s)  & $227202$ & $19344$ & $353874$ & $54$             & $25027$ & $47265$ & $66405$ & $83840$ \tabularnewline
					
					& HN      & NA       & NA      & $1500$  & $42$             & $1278$  & $3521$  & $4253$  & $4751$ \tabularnewline
					
					\hline
					
					& TCR(\%) & $58.88$   & $63.39$ & $53.32*$    & $33.04$ & $64.08$ & $64.54$ & $\textbf{66.99}$ & $65.75$ \tabularnewline
					Forest & SCR(\%) & NA        & NA      & NA    & $64.59$ & $72.65$ & $73.36$ & $\textbf{76.99}$ & $74.32$ \tabularnewline
					Covertype & TrT(s)  & $362354$  & $22878$ & $115946$    & $1249$  & $4730$  & $9766$  & $15002$          & $28597$ \tabularnewline
					& HN      & NA        & NA      & $600$ & $6412$  & $295$   & $311$   & $470$            & $421$   \tabularnewline
					\hline
					& TCR(\%) & $74.87$   & $72.15$ & $51.54*$     & $67.00$  & $75.44$ & $75.77$ & $75.57$ & $\textbf{76.68}$ \tabularnewline
					& SCR(\%) & NA        & NA      & NA     & $70.68$  & $80.11$ & $80.39$ & $79.97$ & $\textbf{81.13}$ \tabularnewline
					SUSY
					& TrT(s)  & $1109396$ & $76387$ & $218441$     & $377$    & $13081$  & $45028$ & $52828$ & $65846$          \tabularnewline
					
					& HN      & NA        & NA      & $1500$ & $111$    & $782$   & $891$   & $1021$  & $954$            \tabularnewline
					\hline 
					& TCR(\%) & $88.13$ & $92.58$ & $49.62$   & $89.55$  & $91.40$ & $92.82$  & $\textbf{93.25}$ & $93.21$ \tabularnewline
					& SCR(\%) & NA      & NA      & NA        & $90.83$  & $92.20$ & $93.17$  & $\textbf{93.50}$ & $93.44$ \tabularnewline
					Hyperplane
					& TrT(s)  & $34094$ & $4810$  & $2480$    & $12$     & $348$   & $622$    & $963$            & $1923$  \tabularnewline
					& HN      & NA      & NA      & $50$      & $11$     & $39$    & $63$     & $76$             & $117$   \tabularnewline
					\hline

					& TCR(\%) & $73.63$ & $76.07$ & $48.84*$    & $81.94$ & $80.24$ & $80.93$ & $82.03$ & $\textbf{83.61}$ \tabularnewline
					Hepmass& SCR(\%) & NA & NA & NA    & $81.52$ & $81.61$ & $81.91$ & $84.57$ & $\textbf{88.24}$ \tabularnewline
					5(\%)& TrT(s)  & $949490$ & $58431$ & $190486$    & $55$    & $4800$  & $6480$  & $10800$ & $15600$           \tabularnewline
					& HN      & NA & NA & $400$ & $45$    & $157$   & $163$   & $170$   & $172$            \tabularnewline
					\bottomrule
					\bottomrule
				\end{tabular}%
				
			}
			\par\end{centering}
		\centering{
			TCR(\%): target classification rate, SCR(\%): source classificarion rate, 
			\\TrT(s): Training time in seconds, HN: hidden nodes, *: Partial results: KDDCup($76.2\%$), \\
			F.Covertype($45.36\%$), Susy($6.94\%$), Hepmass($53.40\%$)  
		}
	\end{table}

	\textbf{Numerical Results}: Conclusive finding can be drawn from Table \ref{result} where ATL  is capable of outperforming its counterparts in five of seven problems with significant reduction of execution time compared to popular multistream algorithms: MSC, FUSION even when being run across 10 epochs. Note that the iteration across epochs here is applied per data chunk rather than the whole data points and thus satisfies the online learning requirement because data chunk is discarded straight away without being seen again in the future. ATL beats its counterpart without any iterative learning in three problems: Forest, SUSY and SEA. ATL is inferior to MSC and FUSION in weather problem but it is caused by the uncertain nature of weather data where NN-based algorithm is always compromised - ADL and TLDA also demonstrate the same behaviour and are worst than ATL. This condition is mainly attributed by the concept of activation function in NN which does not offer any distance information. ATL is on par to MSC in the KDD problem while being slightly ahead of FUSION. 
	
	Although ADL attains the fastest runtime across all problems, it does not implement any domain adaptation algorithm and suffers from accuracy degradation in almost all problems. It is worst than ATL in six of seven problems. The numerical result of ADL in KDD problem should be carefully understood because the covariate shift problem is not apparent in this problem. Opposite situation is observed in the Forest problem where ADL performance is significantly poor while ATL delivers the best performance with the absence of epoch. The Forest problem is well-known for its severe sample bias problem and is always included as a benchmark problem in the literature \cite{FUSION,MSC,MSCRDR}. In addition, ADL is crafted using a deep network structure whereas ATL is built upon a shallow single hidden layer structure. 
	
	TLDA's performance is compromised in the non-stationary problems although it is run with 10 epochs as its original setting but relatively competitive in the stationary problems. The iterative learning of TLDA differs from ATL because the iteration over epoch makes use of all samples and causes substantial increase in the execution time. This aspect is also the underlying reason in only reporting partial results of TLDA in the four large-scale problems: Forest, SUSY, HEPMASS, and KDDCup because we do not successfully produce its results on the paper submission date. ATL characterizes a fully autonomous working principle where the hidden nodes are automatically grown and pruned on the fly while AGMM shares the same trait. The aspect is pictorially demonstrated in the supplemental document.
	\subsection{Source Domain}
	This section outlines ATL's performance in the source domain. This numerical study aims to highlight the underlying spirit of transfer learning across many data streams where a model is supposed to perform equally well in both source and target domain. The numerical study is carried out at the same time as above but instead of measuring the classification performance in the source domain it is undertaken for the target domain. Because FUSION, MSC and TLDA can be evaluated only in the source domain as per their codes, no comparison is performed against them. 
	
	It is observed that ATL still retains decent performance in the source domain and outperforms ADL in six of seven problems. Note that ADL here treats target and source data streams as a single stream and the training process occurs in the source domain while the testing phase is undertaken using the target domain samples. Hence, this finding confirms that the domain adaptation strategy of ATL improves its performance in the source domain and encourages ATL to be domain-invariant for both domains. It is perceived the performance difference in the source and target domains is subtle. 
	
	\subsection{Ablation Study}
	This section elaborates the advantage of ATL's learning components where it is framed under three configurations: (A) The KL divergence training phase is put into a sleep mode; (B) The hidden unit growing and pruning phase is carried out with the absence of AGMM; (C) The hidden unit pruning and growing mechanisms are switched off completely. The SEA dataset is exploited in the ablation study while our numerical study is run twice to arrive at consistent conclusion.
	
	From Table \ref{ablation_result}, the advantage of each learning module is obvious. The absence of KL divergence decreases predictive accuracy in both target and source domains by 1\%. The advantage of AGMM is clearly depicted where removal of AGMM brings down the predictive performance. The absence of AGMM leads ATL to return to conventional one-by-one addition of hidden units deemed to slow to attain desirable network capacity. It is also hindered by too strict assumption of normal distribution and does not have sufficient adaptive mechanism under the concept drift causing current estimation of probability density function to be obsolete. The absence of self-evolving mechanism imposes substantial deterioration on learning performance due to lack of adaptation to concept drifts. Note that the network capacity in (C) is akin to that of the main result in Table \ref{result}. 
	\subsection{Effect of Epochs}
	This section studies the effect of the number of epochs for the ATL's performance. The iteration over a number of epoch occurs per mini batch without revisiting previous mini batches again in the future and complies to the requirement of online learner where the space and time complexity have to be independent to the number of data points. The SGD method is not the exact approximation of closed loop solution and its success is determined by the step size including the number of epochs.  
	
	The number of epoch slightly improves ATL's performance by 2\% margin at most from those of the strictly single pass learning mode. This fact also substantiates the online aspect of ATL where its performance does not depend on the number of epochs and is already comparable to other algorithms without epoch. Despite the increase of execution time, the use of epoch is still much faster than FUSION, MSC and TLDA.

	\begin{table}[htbp]
		\begin{center}

			\caption{Numerical results of ATL's ablation study on SEA dataset}
			\label{ablation_result} \scalebox{1}{
				\begin{tabular}{cc|cc|cc|cc}
					\toprule
					\toprule
					Dataset & \multicolumn{1}{c}{Config} & \multicolumn{2}{c}{A} & \multicolumn{2}{c}{B} & \multicolumn{2}{c}{C} \\
					\midrule
					\midrule
					& TCR (\%) & 90.33 & 90.15 & 90.71 & 90.27 & 90.73 & 90.84 \\
					& SCR ( \%) & 91.98 & 91.69 & 92.11 & 91.97 & 92.47 & 92.30 \\
					SEA   & TrT (s) & 275   & 252   & 101   & 102   & 83    & 81 \\
					& HN    & 44    & 37    & 12    & 13    & 106   & 106 \\
					\bottomrule
					\bottomrule
			\end{tabular}}%
			\\
			\centering{
				TCR(\%): target classification rate, SCR(\%): source classificarion rate, \\
				TrT(s): Training time in seconds, HN: hidden nodes 
			}
		\end{center}
	\end{table}%

	\section{Conclusion}
	Autonomous Transfer Learning (ATL) is proposed to address the transfer learning problem over many streaming processes suffering from no access of ground truth in the target domain, the covariate shift problem as well as the asynchronous drift in the target and source domain. ATL is built upon a domain adaptation technique combining the generative and discriminative phase making use of the noise injecting mechanism of denoising auto encoder (DAE). In addition, the KL divergence learning phase is used to align hidden representation in both domains. All of which aim to address the sample bias problem. The unique property of ATL is perceived in its fully flexible working principle, capable of self-evolving its network structure \textbf{from scratch} in both supervised and unsupervised manner. That is, it is capable of automatically generating its structure based on predictive error and reconstruction error. In addition, AGMM is put forward to increase the flexibility of self-constructing strategy which resolves over-dependence on normal distribution and changing learning environments causing $p(x)_t\neq p(x)_{t+1}$ due to its open structure principle which can grow and shrink its knowledge base on the fly. The hidden unit growing strategy of ATL goes one step ahead of similar works in the literature enabling introduction of many hidden nodes once the hidden unit growing criterion is satisfied. This strategy handles the asynchronous drift issue. Our numerical study in seven problems confirms the advantage of ATL over the baselines in six of seven numerical examples in terms of accuracy and execution time. 
	
	\section*{ACKNOWLEDGEMENT}
	This work is financially supported by the NTU-SUTD-ASTAR AI partnership grant, grant ID: RGANS1902. The fourth author acknowledges the support of the COMET-K2 Center of the Linz Center of Mechatronics (LCM). 
	\linespread{0.75}

	\bibliographystyle{unsrt}  

	\bibliography{references}

\begin{thebibliography}{10}

\bibitem{MSC}
Swarup Chandra, Ahsanul Haque, Latifur Khan, and Charu Aggarwal.
\newblock An adaptive framework for multistream classification.
\newblock In {\em Proceedings of the 25th ACM International on Conference on
  Information and Knowledge Management}, CIKM '16, pages 1181--1190, 2016.

\bibitem{KMM}
Jiayuan Huang, Arthur Gretton, Karsten Borgwardt, Bernhard Sch\"{o}lkopf, and
  Alex~J. Smola.
\newblock Correcting sample selection bias by unlabeled data.
\newblock In {\em Advances in Neural Information Processing Systems 19}, pages
  601--608. 2007.

\bibitem{FUSION}
Ahsanul Haque, Zhuoyi Wang, Swarup Chandra, Bo~Dong, Latifur Khan, and Kevin~W.
  Hamlen.
\newblock Fusion: An online method for multistream classification.
\newblock In {\em Proceedings of the 2017 ACM on Conference on Information and
  Knowledge Management}, CIKM '17, pages 919--928, 2017.

\bibitem{KLIEP}
Masashi Sugiyama, Shinichi Nakajima, Hisashi Kashima, Paul~V. Buenau, and
  Motoaki Kawanabe.
\newblock Direct importance estimation with model selection and its application
  to covariate shift adaptation.
\newblock In {\em Advances in Neural Information Processing Systems 20}, pages
  1433--1440. 2008.

\bibitem{PEARSON}
Makoto Yamada, Taiji Suzuki, Takafumi Kanamori, Hirotaka Hachiya, and Masashi
  Sugiyama.
\newblock Relative density-ratio estimation for robust distribution comparison.
\newblock In {\em Advances in Neural Information Processing Systems 24}, pages
  594--602. 2011.

\bibitem{ADL}
Andri Ashfahani and Mahardhika Pratama.
\newblock Autonomous deep learning: Continual learning approach for dynamic
  environments.
\newblock In {\em In SIAM International Conference on Data Mining}, 2019.

\bibitem{DRCN}
Muhammad Ghifary, W.~Bastiaan Kleijn, Mengjie Zhang, David Balduzzi, and Wen
  Li.
\newblock Deep reconstruction-classification networks for unsupervised domain
  adaptation.
\newblock In {\em Computer Vision - {ECCV} 2016 - 14th European Conference,
  Amsterdam, The Netherlands, October 11-14, 2016, Proceedings, Part {IV}},
  pages 597--613, 2016.

\bibitem{TLDA}
Fuzhen Zhuang, Xiaohu Cheng, Ping Luo, Sinno~Jialin Pan, and Qing He.
\newblock Supervised representation learning: Transfer learning with deep
  autoencoders.
\newblock In {\em Proceedings of the 24th International Conference on
  Artificial Intelligence}, IJCAI'15, pages 4119--4125, 2015.

\bibitem{learning_dynamic_AE}
Arnu Pretorius, Steve Kroon, and Herman Kamper.
\newblock Learning dynamics of linear denoising autoencoders.
\newblock In {\em Proceedings of the 35th International Conference on Machine
  Learning}, volume~80 of {\em Proceedings of Machine Learning Research}, pages
  4141--4150, Stockholmsmässan, Stockholm Sweden, 10--15 Jul 2018.

\bibitem{VincentDAE}
Pascal Vincent, Hugo Larochelle, Yoshua Bengio, and Pierre-Antoine Manzagol.
\newblock Extracting and composing robust features with denoising autoencoders.
\newblock In {\em Proceedings of the 25th International Conference on Machine
  Learning}, ICML '08, pages 1096--1103, New York, NY, USA, 2008. ACM.

\bibitem{Murphy_Machine_Learning}
Kevin~P. Murphy.
\newblock {\em Machine Learning: A Probabilistic Perspective}.
\newblock The MIT Press, 2012.

\bibitem{GamaDataStream}
Joao Gama.
\newblock {\em Knowledge Discovery from Data Streams}.
\newblock Chapman \& Hall/CRC, 1st edition, 2010.

\bibitem{BARTFIS}
Richard~Jayadi Oentaryo, Meng~Joo Er, Linn San, and Xiang Li.
\newblock Online probabilistic learning for fuzzy inference system.
\newblock {\em Expert Syst. Appl.}, 2014.

\bibitem{forestcover}
Mohammad Masud, Jing Gao, Latifur Khan, Jiawei Han, and Bhavani~M.
  Thuraisingham.
\newblock Classification and novel class detection in concept-drifting data
  streams under time constraints.
\newblock {\em IEEE Trans. on Knowl. and Data Eng.}, 23(6):859--874, June 2011.

\bibitem{SEA}
W.~N. Street and Y-S Kim.
\newblock A streaming ensemble algorithm (sea) for large-scale classification.
\newblock In {\em Proceedings of the Seventh ACM SIGKDD International
  Conference on Knowledge Discovery and Data Mining}, KDD '01, pages 377--382,
  2001.

\bibitem{Baldi2014SearchingFE}
P.~Baldi, Paul~D. Sadowski, and Daniel Whiteson.
\newblock Searching for exotic particles in high-energy physics with deep
  learning.
\newblock {\em Nature communications}, 5:4308, 2014.

\bibitem{KDDCup}
Salvatore~J. Stolfo, Wei Fan, Wenke Lee, Andreas Prodromidis, and Philip~K.
  Chan.
\newblock Cost-based modeling for fraud and intrusion detection: Results from
  the jam project.
\newblock In {\em In Proceedings of the 2000 DARPA Information Survivability
  Conference and Exposition}, pages 130--144. IEEE Computer Press, 2000.

\bibitem{Bifet07learningfrom}
Albert Bifet and Ricard Gavaldà.
\newblock Learning from time-changing data with adaptive windowing.
\newblock In {\em In SIAM International Conference on Data Mining}, 2007.

\bibitem{DitzlerImbalanced}
G.~Ditzler and R.~Polikar.
\newblock Incremental learning of concept drift from streaming imbalanced data.
\newblock {\em IEEE Trans. on Knowl. and Data Eng.}, 25(10):2283--2301, October
  2013.

\bibitem{datastreamevaluation}
Jo{\~{a}}o Gama, Raquel Sebasti{\~{a}}o, and Pedro~Pereira Rodrigues.
\newblock On evaluating stream learning algorithms.
\newblock {\em Machine Learning}, 90(3):317--346, 2013.

\bibitem{MSCRDR}
Bo~Dong, Swarup Chandra, Yang Gao, and Latifur Khan.
\newblock Multistream classification with relative density ratio estimation.
\newblock In {\em AAAI 2019}, 2019.

\end{thebibliography}

\end{document}